\documentclass[runningheads]{llncs}
\usepackage[T1]{fontenc}
\usepackage{graphicx}
\usepackage{float}
\usepackage{multirow}
\usepackage{booktabs}
\usepackage{threeparttable}
\usepackage{supertabular}
\usepackage{ulem}
\usepackage{subfigure}
\usepackage{graphicx}
\usepackage{comment}
\begin{document}
\title{Speeding Up Question Answering Task of Language Models via Inverted Index}
%
%
\author{
  Xiang Ji
  \and
  Yesim Sungu-Eryilmaz
  \and
  Elaheh Momeni
  \and
  Reza Rawassizadeh\\
  \texttt{ jjixiang@bu.edu, yesims@bu.edu, momeni.elaheh@gmail.com,rezar@bu.edu}
}

\authorrunning{Xiang et al.}
%
\institute{Boston University, Boston MA 02215, USA \and
eMentalist, Vienna, Austria
}
\maketitle            

\begin{abstract}
Natural language processing applications, such as conversational agents and their question-answering capabilities, are widely used in the real world. Despite the wide popularity of large language models (LLMs), few real-world conversational agents take advantage of LLMs. Extensive resources consumed by LLMs disable developers from integrating them into end-user applications. In this study, we leverage an inverted indexing mechanism combined with LLMs to improve the efficiency of question-answering models for closed-domain questions. Our experiments show that using the index improves the average response time by 97.44\%. In addition, due to the reduced search scope, the average BLEU score improved  by 0.23 while using the inverted index.

\end{abstract}
\vspace{-1cm}
\section{Introduction and background}
Advances in large language models (LLMs), especially employing the transformer architecture \cite{BERT}, revolutionized the quality of natural language processing applications. However, training an LLM with state-of-the-art architecture, including several layers of decoder/encoder and attention, is computationally very expensive and not possible by small and medium enterprises with limited budgets. 

There are promising works to reduce the model sizes of LLMs such as compression \cite{holmes2022compressing,tao2022compression}, quantization \cite{yu2020neural} or the use of knowledge distillation \cite{Computation}, but still, larger models are favored due to their accuracy. 

This issue leads to the introduction of services such as huggingface\cite{hugging} or spacy\cite{spacy2}, which can share the trained (pre-trained and fine-tuned) LLMs. These services enable developers to benefit from trained LLM, while not dealing with the expensive model training process. However, executing a query on trained LLMs still consumes a significant amount of time. As the number of words in the text increases, the response time of these models in answering the questions increases exponentially. For example, if LLMs are used to answer questions based on a book, they take an extremely long time to answer a question, which is not practical in real-world applications. Response time has a direct correlation with the usability of the application \cite{SchuetzlerGG18,rong2022odsearch}. Therefore, slow response time hinders the adaption of large language models into end-user applications. 

In this work, we employ five popular transformer models for question answering (Q\&A), i.e. BERT-base\cite{BERT}, BERT-large\cite{BERT}, DistilBERT\cite{distailBERT}, RoBERTa\cite{robert}, and Tiny-RoBERTa\cite{robert}. Then, we develop an inverted index layer \cite{Zobel2006InvertedFF} that reduces the execution time while maintaining and improving the accuracy of the Q\&A models. While extracting answers from a large amount of text (such as a book), our result shows that our approach has a significant improvement in both response time and accuracy over the baseline models (not using an inverted index). 
\vspace{-0.3cm}
\section{Method}
\begin{figure}[H]
     \vspace{-0.5cm}
    \centering
    \includegraphics[scale=0.30]{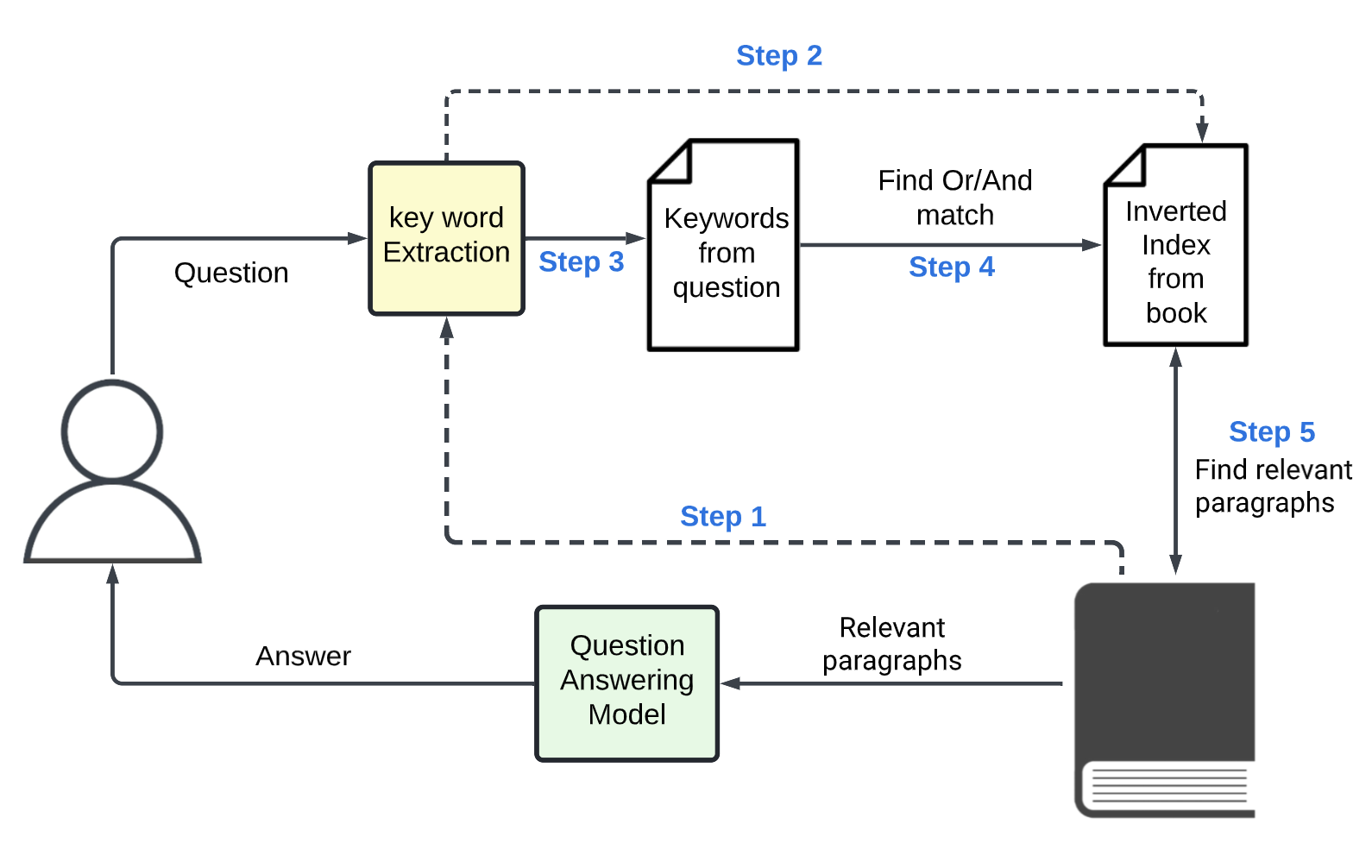}
    \vspace{-3mm}
    \caption{Our proposed architecture. First, it identifies the keywords from the user's question and then searches the relevant paragraphs based on extracted user keywords. Dotted lines present phase 1 and straight lines present phase 2.}
    \label{fig:arch}
    \vspace{-0.5cm}
\end{figure}

\vspace{-0.3cm}
Our proposed architecture (see Figure \ref{fig:arch})  includes two phases and five steps. Phase 1 includes two steps. Step 1 analyzes the large corpus of the text and extracts its keywords from each paragraph. For example, the keyword “bench press“ was extracted from several paragraphs from a fitness book, and  "SVM" was extracted from several paragraphs from a machine learning book. 
\vspace{-3mm}

Step 2 builds an inverted index on the extracted keywords on the paragraph level for the given book.
Phase 1 (steps 1 and 2) should be done offline and once for every book. To build the inverted index, we do not use an existing library such as Lucene \cite{lucene} or SOLR \cite{solr}. Instead, we perform the indexing locally to have control over the accuracy of extracted keywords, which is impossible to do with the current inverted indexing method. 

In Phase 2, step 3 parses the user's question and extracts its keywords. Step 4 matches the keywords from the question with keywords inside the inverted index. Step 5, which checks if there is a match, then it finds the given page and paragraph of the index and feeds this text into the LLM module (e.g. BERT) to find the answer. Instead of searching the entire text corpus, it only searches locations specified by an inverted index.

To implement Phase 1, step 1, we experiment with two keyword extraction algorithms (Rake\cite{rake} and keyBERT\cite{grootendorst2020keybert}) to extract one or two keywords for each paragraph of the book in paragraph order. We choose keyBERT for keyword extraction since it outperforms Rake. 

The result of keyword extraction and the source of keyword extraction is stored in a table, which constitutes the inverted index table. The following presents three examples of invert indexes created on “Fitness Mindset” by Brian Keane \cite{fitbook}.\\
{\fontfamily{qcr}\selectfont
page num:10, paragraph num:1,4, keyword: losing weight \\
page num:15, paragraph num:2, keyword: hormone interact \\
page num:12, paragraph num:1, keyword: strength conditioning \\
}
Next, when a user asks a question, our system extracts the keywords from the user's question (see Figure \ref{fig:arch}, yellow section). For example, the user asks: \textit{“What are the effects of eating sugar?”}. The extracted keywords by KeyBERT are “effect”, “eat” and “sugar”. Afterward, keywords extracted from user queries will be matched with the keywords in the inverted index (they have been constructed in phase 1). Since the inverted index includes the location of related paragraphs and pages in the book, there is no need for the LLM to search the whole book for finding the answer. Instead, it only searches related paragraphs.
\vspace{-0.3cm}
\section{Experiment}
\vspace{-0.3cm}
\subsection{Data Corpus}
We chose three books from three distinct disciplines—fitness, fiction, and computer science for our experiments. The first book is  “Fitness Mindset” by Brian Keane\cite{fitbook}. The book has 174 pages (28,732 words); we refer to it as Book A.
The second book is the “Count of Monte Cristo”\cite{montebook}, a total of 1342 pages (464,162 words); we refer to it as Book B. The third book is an in-progress machine learning book, i.e., “Concepts and Algorithms of Machine Learning” \cite{MLBOOK}. At the time of writing this paper, approximately 810 pages are written, we refer to this book as Book C.
   \vspace{-3mm}
\begin{table}[htb]
   \addtolength{\leftskip} {-1cm}
    \caption{Compare two methods of extracting keywords, Rake and keyBERT}
    \begin{tabular}{|c|c|c|}
    \hline
         Content&Rake&keyBERT  \\\hline
         \shortstack{how many calories do you need this number \\ can vary greatly from person to person.}& [('vary greatly', 4.0)]&[('calories need', 0.7308)]\\\hline
         \shortstack{note the same formula can be applied to any food \\item to calculate the number of calories.l}&[('food item', 4.0)]&[('number calories', 0.6844)]\\\hline
\shortstack{in addition at the most basic level if you eat more \\calories than you burn you will add weight.}&[('basic level', 4.0)]&[('add weight', 0.5402)]\\\hline
    \end{tabular}
    \label{tab:keywordext}
\end{table}
\vspace{-9mm}
\subsection{Keyword Extraction}
\vspace{-3mm}
Our first experiment focuses on selecting the best keyword extraction method. We choose to compare Rake and keyBERT. As an example, three extracted keywords from each sentence of Brian Keane's "Fitness Mindset" \cite{fitbook} are shown in Table \ref{tab:keywordext}. 

After experimenting with 50 sentences, two human experts analyzed the results and  found that keyBERT is 26\% more accurate than Rake's result. The Fleiss' Kappa of their agreement is 0.85 which is almost perfect agreement. Therefore, we use KeyBERT to extract keywords from the text and build the inverted index. The index creation for a given book is done once, and then indexes are stored in the index table. 
\vspace{-8mm}
\begin{figure}[H]
    \centering
    \hspace*{-2cm}
    \includegraphics[scale=0.58]{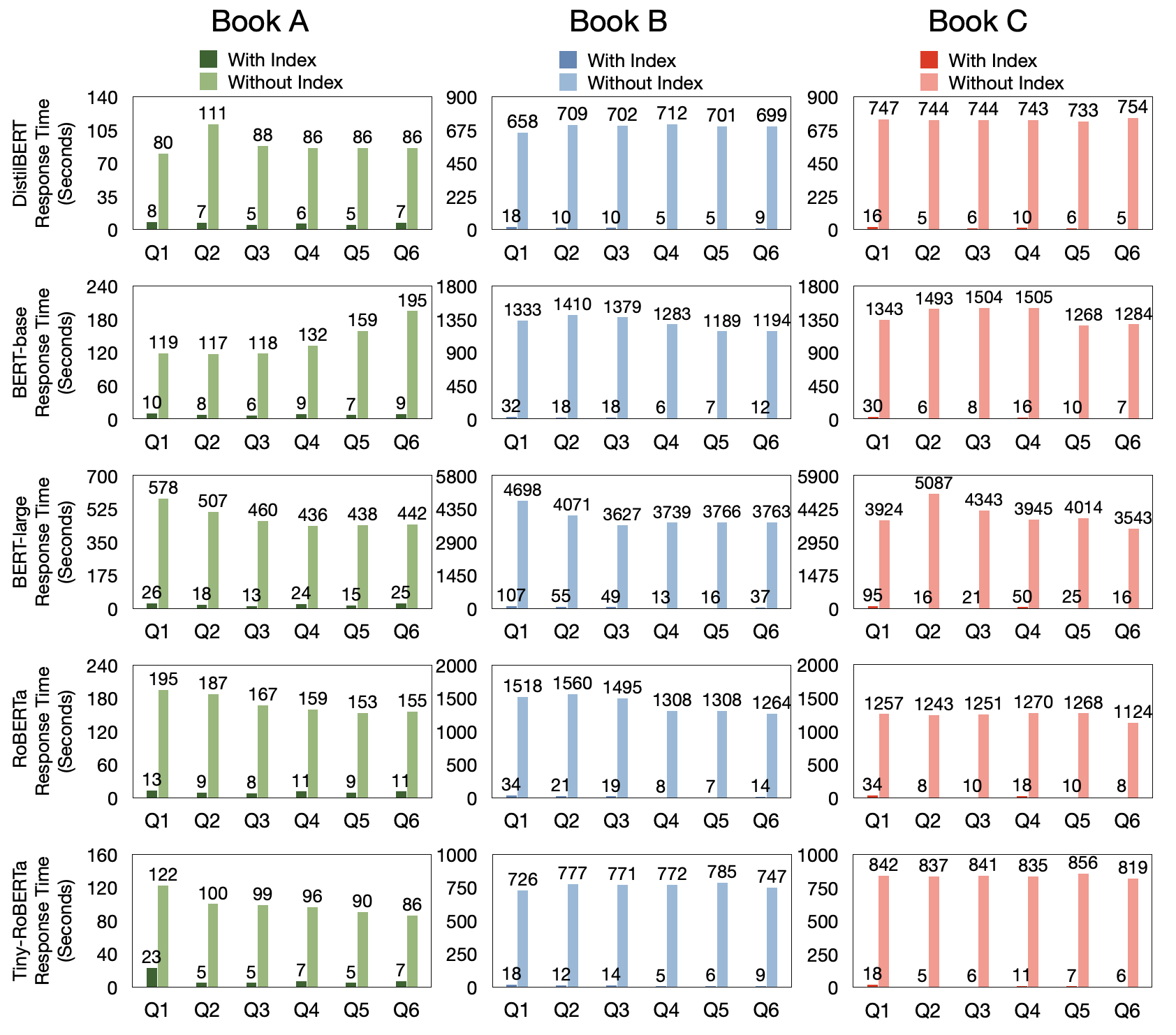}
    \caption{Response time comparison with book A, B and C, between five selected language models.}
    \label{fig:restime}
\end{figure}
\vspace{-8mm}
\subsection{Response Time }
The second experiment focuses on exploring the difference in query time by using our proposed indexing versus not using the proposed indexing method. 

We have constructed six questions for each book. Questions are related to the content of each book, and the answer exists inside the book. Even if the answer does not exist, our approach is faster than the baseline (using the LLM without an inverted index) because instead of searching the entire book, the algorithm only searches for keywords inside the inverted index. The list of questions can be found in the source code \cite{Ji_LargeBook_2022}. 
To conduct the response time analysis, we used a brute force approach, which feeds the entire book into the LLM, versus using our inverted index approach. The experiment was conducted on three books and six questions for each book. The result of the experiment is shown in Figure \ref{fig:restime}.

Our approach uses about 10.7 seconds on average to answer the query, and the baseline uses 194.9 seconds to answer the query. This shows that using an inverted index improves 94.51\% of the query execution time for book A. Our approach uses about 19.8 seconds on average to answer the query, and the baseline uses 1622.13 seconds to answer the query. This shows that using an inverted index improves 98.77\% the query execution time for book B. Our approach uses about 16.3 seconds on average to answer the query, and the baseline uses 1,672 seconds to answer the query. This shows that using an inverted index improves 99.02\% of the query execution time for book C.

\subsection{Experiments with quality of answers}
Since the inverted index significantly reduces the search space for LLM, we investigate its impact on the accuracy of the Q\&A task as well. In this experiment, we use BLEU score \cite{papineni-etal-2002-bleu} to compare the result of using the inverted index versus not using the inverted index.

To perform this experiment, we used ten questions for each book, and two human annotators manually provided standard answers. Afterward, we asked these questions, once using the inverted index and once not using it (baseline). Next, we calculate the average of the BLEU scores\cite{papineni-etal-2002-bleu} for both cases. The higher the BLEU score, the more similar the model response is to the human response. The followings are  question-answer examples and their BLEU scores\cite{papineni-etal-2002-bleu}. \\
{\fontfamily{qcr}\selectfont
Q: "Book C- What is a hidden layer?".\\
A: “layers between the input layer and output layers”\\
BLEU score: 1 \\
Q: "Book C- What is the use of PCA?".\\
A: “transforms the original dataset into a new dataset”\\
BLEU score: 0.76 \\
Q: "Book A- What is the sugar effect?". \\
A: "Since then" \\
BLEU score: 0 \\
}

Each model (with and without the inverted index) will get the answer. We can measure the quality of the model answers by comparing the BLEU score\cite{papineni-etal-2002-bleu}.

\begin{figure}[H]
    \hspace*{-1cm}
    \centering
    \includegraphics[scale=0.45]{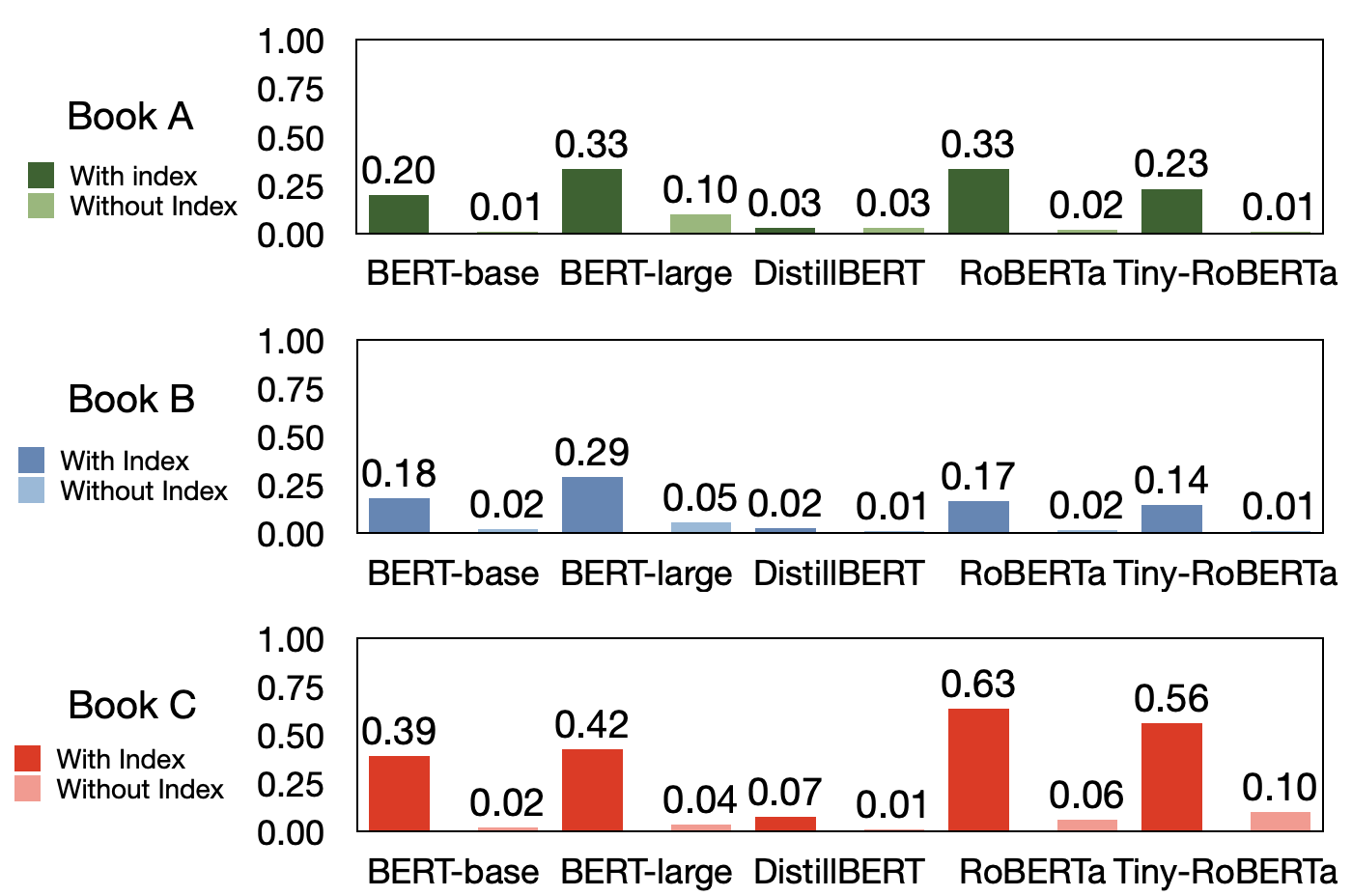}
    \caption{BLEU Score Comparison between using and not using the inverted index. As is shown in the result, using the inverted index significantly BLEU score of results.}  
    \label{fig:BLEU}
\end{figure}

The results of comparisons are presented in Figure \ref{fig:BLEU}.

The result of the experiment presents the average BLEU score of 0.034 for book A, before using the index method, but using the index method, the average BLEU score was 0.224. Before using the inverted index, the average BLEU score is 0.022 for book B, but after using the inverted index, the average BLEU score was 0.16. Before using the inverted index, the average BLEU score was 0.05 for book C, but after using the index method, the average BLEU score was 0.41. The BLEU score was improved by 0.19, 0.14, and 0.36 for books A, B, and C. This increase demonstrates the significant impact of using the inverted index on the accuracy of results, i.e. BLEU score increase by 0.23.

\section{Conclusion}
In this work, we tackle the response time problem of trained LLMs, that hinders their proliferation into end-user applications. Using an inverted index, we demonstrate that both the response time and accuracy of the trained Q\&A models increase. The augmentation we proposed can enable end users' applications to integrate Q\&A models to answer closed-domain questions, even on devices with limited computational capabilities, such as smartphones and wearable devices.
\bibliographystyle{splncs04}
\bibliography{mybibfile.bib}

\end{document}